\documentclass[journal]{IEEEtran}
\usepackage{bm}

\newcommand{\NumberOfBands}{b}
\newcommand{\DiagonalMatrixNoise}{\Sigma}

\newcommand{\NormalDistribution}{\mathcal{N}}
\newcommand{\TestTimeGeneratedSamples}{A}

\usepackage{amsfonts}
\usepackage{graphicx}
\usepackage{makecell}
\usepackage{tikz}
\usepackage{collcell}
\usepackage{colortbl}
\usepackage{pgfplots}
\usepackage{pgfplotstable}
\usepackage{multirow}
\usepackage{url}
\usepackage[all]{xy}
\usepackage{cite}
\usepackage{booktabs}
\usepackage{adjustbox}
\usepackage{tikz}
\usetikzlibrary{arrows,positioning}
\usetikzlibrary{plotmarks}
\usetikzlibrary{calc}

\newcommand{\TrainingSample}{\bm{t}} 
\newcommand{\InputSample}{\bm{x}} 
\newcommand{\DataMatrix}{\mathcal{D}}
\newcommand{\CovarianceMatrix}{\mathcal{C}}

\newcommand{\term}[1]{\mathfrak{#1}}
\newcommand{\zbior}[1]{\mathbb{#1}}  
\newcommand{\macierz}[1]{{\boldsymbol{\mathrm{#1}}}}
\newcommand{\wektor}[1]{\macierz{\MakeLowercase{#1}}}
\newcommand{\newItem}[2]{%
  \expandafter\def\csname #1\endcsname {\MakeLowercase{#2}} %
  \expandafter\def\csname n#1\endcsname {\MakeUppercase{#2}} %
  \expandafter\def\csname zb#1\endcsname {\zbior{\MakeUppercase{#2}}} %
  \expandafter\def\csname t#1\endcsname {\term{\MakeLowercase{#2}}} %
  \expandafter\def\csname m#1\endcsname {\macierz{\MakeUppercase{#2}}} %
  \expandafter\def\csname w#1\endcsname {\wektor{#2}} %
  }
  \newItem{Atrybut}{d_f}
\newItem{Deskryptor}{x}
\newItem{Regula}{l}
\newItem{Przeslanka}{a}
\newItem{Konsekwencja}{b}
\newItem{Implikacja}{b'}
\newItem{Zespolona}{c}
\newItem{Rzeczywista}{r}
\newItem{Klasa}{\delta}
\newItem{Decyzja}{y}
\newItem{Przyklad}{a}
\newItem{Test}{a}
\newItem{Klaster}{c}
\newItem{Rozmycie}{s}
\newItem{Centre}{v}
\newItem{Membership}{u}
\newItem{ClusterMembership}{\mu}
\newItem{Weight}{z}
\newItem{Krotka}{x}
\newItem{Width}{w}
\newItem{Wspolczynnik}{p}
\newItem{Obiekt}{k}
\newItem{Odleglosc}{t}
\newItem{Parameter}{p}

\newcommand{\TrainingSet}{\bm{T}} 
\newcommand{\TrainingSetSize}{N}
\newcommand{\ValidationSet}{\bm{V}} 

\newcommand{\TestSet}{\Psi} 

\ifCLASSINFOpdf
\else
\fi

\pgfplotstableset{
    /color cells/min/.initial=0,
    /color cells/max/.initial=1000,
    /color cells/textcolor/.initial=,
    color cells/.code={%
        \pgfqkeys{/color cells}{#1}%
        \pgfkeysalso{%
            postproc cell content/.code={%
                \begingroup
                \pgfkeysgetvalue{/pgfplots/table/@preprocessed cell content}\value
\ifx\value\empty
\endgroup
\else
                \pgfmathfloatparsenumber{\value}%
                \pgfmathfloattofixed{\pgfmathresult}%
                \let\value=\pgfmathresult
                \pgfplotscolormapaccess
                    [\pgfkeysvalueof{/color cells/min}:\pgfkeysvalueof{/color cells/max}]%
                    {\value}%
                    {\pgfkeysvalueof{/pgfplots/colormap name}}%
                \pgfkeysgetvalue{/pgfplots/table/@cell content}\typesetvalue
                \pgfkeysgetvalue{/color cells/textcolor}\textcolorvalue
                \toks0=\expandafter{\typesetvalue}%
                \xdef\temp{%
                    \noexpand\pgfkeysalso{%
                        @cell content={%
                            \noexpand\cellcolor[rgb]{\pgfmathresult}%
                            \noexpand\definecolor{mapped color}{rgb}{\pgfmathresult}%
                            \ifx\textcolorvalue\empty
                            \else
                                \noexpand\color{\textcolorvalue}%
                            \fi
                            \the\toks0 %
                        }%
                    }%
                }%
                \endgroup
                \temp
\fi
            }%
        }%
    }
}

\begin{document}
\title{Hyperspectral Data Augmentation}
\author{Jakub Nalepa,~\IEEEmembership{Member,~IEEE}, 
        Michal Myller,
        and Michal Kawulok,~\IEEEmembership{Member,~IEEE}
\thanks{This work was funded by European Space Agency (HYPERNET project).}
\thanks{J.~Nalepa, M.~Myller, and M.~Kawulok are with Silesian University of Technology, Gliwice, Poland (e-mail: \{jnalepa, michal.kawulok\}@ieee.org), and KP Labs, Gliwice, Poland (\{jnalepa, mmyller, mkawulok\}@kplabs.pl).}
}


\maketitle
\begin{abstract}
Data augmentation is a popular technique which helps improve generalization capabilities of deep neural networks. It plays a pivotal role in remote-sensing scenarios in which the amount of high-quality ground truth data is limited, and acquiring new examples is costly or impossible. This is a common problem in hyperspectral imaging, where manual annotation of image data is difficult, expensive, and prone to human bias. In this letter, we propose online data augmentation of hyperspectral data which is executed during the inference rather than before the training of deep networks. This is in contrast to all other state-of-the-art hyperspectral augmentation algorithms which increase the size (and representativeness) of training sets. Additionally, we introduce a new principal component analysis based augmentation. The experiments revealed that our data augmentation algorithms improve generalization of deep networks, work in real-time, and the online approach can be effectively combined with offline techniques to enhance the classification accuracy.
\end{abstract}


\IEEEpeerreviewmaketitle

\section{Introduction} \label{sec:intro}

Hyperspectral satellite imaging (HSI) captures a wide spectrum (commonly more than a hundred of bands) of light per pixel (forming an array of reflectance values). Such detailed information is being exploited by the remote sensing, pattern recognition, and machine learning communities in the context of accurate HSI \emph{classification} (elaborating a class label of an HSI pixel) and \emph{segmentation} (finding the boundaries of objects) in many fields~\cite{Dundar2018}. Although the segmentation techniques include conventional machine learning algorithms (both unsupervised~\cite{Bilgin2011} and supervised~\cite{Dundar2018,Li2018}), deep learning based techniques became the main stream~\cite{Chen2015,Zhao2016,Chen2016,Zhong2017,Mou2017,Santara2017,Lee2017,Gao_2018}. They encompass deep belief networks~\cite{Chen2015,Zhong2017}, recurrent neural networks~\cite{Mou2017}, and convolutional neural networks (CNNs)~\cite{Zhao2016,Chen2016,Santara2017,Lee2017,Gao_2018}.

Deep neural networks discover the underlying data representation, hence they do not require feature engineering and can potentially capture features which are unknown for humans. However, to train such large-capacity learners (and to avoid overfitting), we need huge amount of ground truth data. Acquiring such training sets is often expensive, time-consuming, and human-dependent. These problems are very important real-world obstacles in training well-generalizing models (and validating such learners) faced by the remote sensing community---they are manifested by a very small number of ground truth benchmark HSI sets (there are approx. 10 widely-used benchmarks, with the Salinas Valley, Pavia University, and Indian Pines scenes being the most popular).

To combat the problem of limited, non-representative, and imbalanced training sets, \emph{data augmentation} can be employed. It is a process of synthesizing new examples following the original data distribution. Since such enhanced training sets can improve generalization of the learners, data augmentation may be perceived as implicit regularization. In computer vision tasks, data augmentation often involves simple \emph{affine} (e.g.,~rotation or scaling) and \emph{elastic} transforms of the image data~\cite{Kri2012ImageNet}. These techniques, albeit applicable to HSI, do not benefit from all available information to model useful data.

\subsection{Related literature}

The literature on HSI data augmentation is fairly limited (not to mention, only \emph{one} of the deep-learning HSI segmentation methods discussed earlier in this section used augmentation---simple mirroring of training samples---for improved classification~\cite{Lee2017}). In~\cite{Slavkovikj:2015:HIC:2733373.2806306}, the authors calculated per-spectral-band standard deviation (for each class) in the training set. The augmented samples are later drawn from a zero-mean multivariate normal distribution $\NormalDistribution(0,\alpha\DiagonalMatrixNoise)$, where $\DiagonalMatrixNoise$ is a diagonal matrix with the standard deviations (for all classes) along its main diagonal, and $\alpha$ is a hyper-parameter (scale factor) of this technique. Albeit its simplicity, this augmentation was shown to be able to help improve generalization.

Li et al. utilized both spectral and spatial information to synthesize new samples in their pixel-block augmentation~\cite{Li2018GRSL}. Two data-generation approaches: (i)~Gaussian smoothing filtering alongside (ii)~label-based augmentation were exploited in~\cite{Acquarelli2018}. The latter technique resembles weak-labeling~\cite{Sun:2010:MLW:2898607.2898703}, and builds on an assumption that neighboring HSI pixels should share the same class (the label of a pixel propagates to its neighbors, and these generated examples are inserted into the training set). Thus, it may introduce wrongly-labeled samples.


Generative adversarial networks (GANs) have already attracted research attention in the context of data augmentation due to their ability of introducing invariance of models with respect to affine and appearance variations. GANs model an unknown data distribution based on the provided samples, and they are composed of a \emph{generator} and \emph{discriminator}. A generator should generate data samples which follow the underlying data distribution and are indistinguishable from the original data by the discriminator (hence, they compete with each other). In a recent work~\cite{audebert2018generative}, Audebert et al.~applied GAN conditioning to ensure that the synthesized HSI examples (from random distribution) belong to the specified class. Overall, all of the state-of-the-art HSI augmentation methods are aimed at increasing the size and representativeness of training sets which are later fed to train the deep learners. 

\subsection{Contribution}

In this letter, we propose a novel \emph{online augmentation} technique for hyperspectral data (Section~\ref{sub:online})---instead of synthesizing samples and adding them to the training set (hence increasing its size which adversely affects the training time of deep learners), we generate new examples \emph{during the inference}. These examples (both original and artificial) are classified using a deep net trained over the original set, and we apply the voting scheme to elaborate the final label. To our knowledge, such online augmentation has not been exploited in HSI analysis so far (test-time augmentation was utilized in medical imaging~\cite{testtimeaugmentation2018}, where the authors applied affine transforms and noise injection into brain-tumor images for better segmentation). Also, we introduce principal component analysis (PCA) based augmentation (Section~\ref{sub:PCA}) which may be used both \emph{offline} (before the training) and \emph{online}. This PCA-based augmentation simulates data variability, yet follows the original data distribution (which GANs are intended to learn~\cite{audebert2018generative}, but they are not applicable at test-time).

Our rigorous experiments performed over HSI benchmarks revealed that the online approach is very flexible---different augmentation techniques can be exploited in this setting (Section~\ref{sec:experiments}). The results obtained using a spectral CNN indicated that the test-time augmentation significantly improves abilities of the models when compared with those trained using the original sets, and augmented using other algorithms (also, we compared our CNN with a spectral-spatial CNN from the literature whose capacity is much larger~\cite{Gao_2018}). Our online approach does not sacrifice the inference speed and allows for real-time classification. We showed that the proposed PCA augmentation is extremely fast, and ensures the highest-quality generalization of the deep models for all data-split scenarios. Finally, we demonstrated that the offline and online augmentations can be effectively combined for better classification.

\section{Proposed Hyperspectral Data Augmentation}\label{sec:method}

\subsection{Online Hyperspectral Data Augmentation}\label{sub:online}

Our online (test-time) data augmentation involves synthesizing $\TestTimeGeneratedSamples$ artificial samples $\InputSample'_1,\InputSample'_2,\dots,\InputSample'_{\TestTimeGeneratedSamples}$ for each incoming example $\InputSample$ during the inference. We traverse the neighborhood of the original example and try to mitigate potential input-dependent uncertainty of the deep model. In contrast to the offline augmentation techniques, the test-time augmentation does not cause increasing the training time of the network, and it does not require defining the number of synthesized samples beforehand (also, for the majority of specific augmentation algorithms, the operation time of a trained learner would not be significantly affected since the inference is fast). We build upon the theory of ensemble learning, where elaborating a combined classifier (encompassing several weak learners) delivers high-quality generalization (it is an efficient regularizer). Here, by creating $\TestTimeGeneratedSamples$ artificial data points, we implicitly form a homogeneous ensemble of deep models (trained over the same training set $\TrainingSet$). The final class label is elaborated using \emph{majority voting} (with equal weights) over all ($\TestTimeGeneratedSamples+1$) samples (for low ensemble confidence, when two or more classes receive the same number of votes, we perform \emph{soft voting}---we average all class probabilities, and the final class label corresponds to the class with the highest average probability).

The proposed online HSI augmentation may be considered to be a meta-algorithm, in which a specific data augmentation method is applied to synthesize samples on the fly. Although we exploited the noise injection based approach~\cite{Slavkovikj:2015:HIC:2733373.2806306}, and our principal component analysis based technique (see Section~\ref{sub:PCA}) in this work, we anticipate that other augmentations which are aimed at modifying an existent sample can be straightforwardly utilized here. Finally, the online augmentation may be coupled with any offline technique (Section~\ref{sec:experiments}).

\subsection{Principal component analysis based data augmentation}
\label{sub:PCA}
In this work, we propose a new augmentation method based on PCA. Let us consider a training set $\TrainingSet$ of $\TrainingSetSize$ HSI pixels $\TrainingSample_i$, where $i=1,2,\dots,\TrainingSetSize$, and each $\TrainingSample_i$ is $\NumberOfBands$-dimensional ($\NumberOfBands$ denotes the number of bands in this HSI). PCA extracts a set of $\NumberOfBands'$ ($\NumberOfBands'\ll\NumberOfBands$) projection directions (vectors) by maximizing the projected variance of a given $\NumberOfBands$-dimensional dataset---the first principal component ($PC_1$) accounts for as much of the data variability as possible, and so forth. First, we center the data at the origin (hence, we subtract the average sample $\bar{\TrainingSample}=\sum_{i=1}^{\TrainingSetSize}\TrainingSample_i/\TrainingSetSize$ from each $\TrainingSample_i$, and form the data matrix $\DataMatrix$ (of $\NumberOfBands\times\TrainingSetSize$ size), whose $i$th column is ($\TrainingSample_i-\bar{\TrainingSample}$). The $b\times b$ covariance matrix becomes $\CovarianceMatrix=(1/\TrainingSetSize)\DataMatrix\DataMatrix^{\rm T}$, and it undergoes eigendecomposition $\CovarianceMatrix=\bm{\Phi}\bm{\Lambda}\bm{\Phi}^T$, where $\bm{\Lambda}=diag(\lambda_1,\lambda_2,\dots,\lambda_{b})$ is the matrix with the non-increasingly ordered eigenvalues along its main diagonal, and $\bm{\Phi}=\left[PC_1|PC_2|\dots|PC_{b}\right]$ is the matrix with $b$ corresponding eigenvectors (principal components) as columns. Finally, $\NumberOfBands'$ principal components form an orthogonal base, and each sample $\TrainingSample$ can be projected onto a new feature space: $\TrainingSample'=\bm{\Phi}^T\TrainingSample$. Importantly, each sample $\TrainingSample'$ can be projected back to its original space: $\TrainingSample''=\bm{\Phi}\TrainingSample'$ with the error $\epsilon=\sum_{i=1}^{\NumberOfBands}(t_i-t_i^{''})$ (the PCA-training procedure minimizes this error---it is non-zero if $\NumberOfBands'<b$; otherwise, if $\NumberOfBands'=b$, there is no reconstruction error).

\begin{figure}[ht!]
\centering
\newcommand{\mywidth}{0.33}
\hspace*{-0.1cm}
\setlength\tabcolsep{1pt}
\begin{tabular}{cccc}
(a) & (b) & (c)\\
  \includegraphics[width=\mywidth\columnwidth]{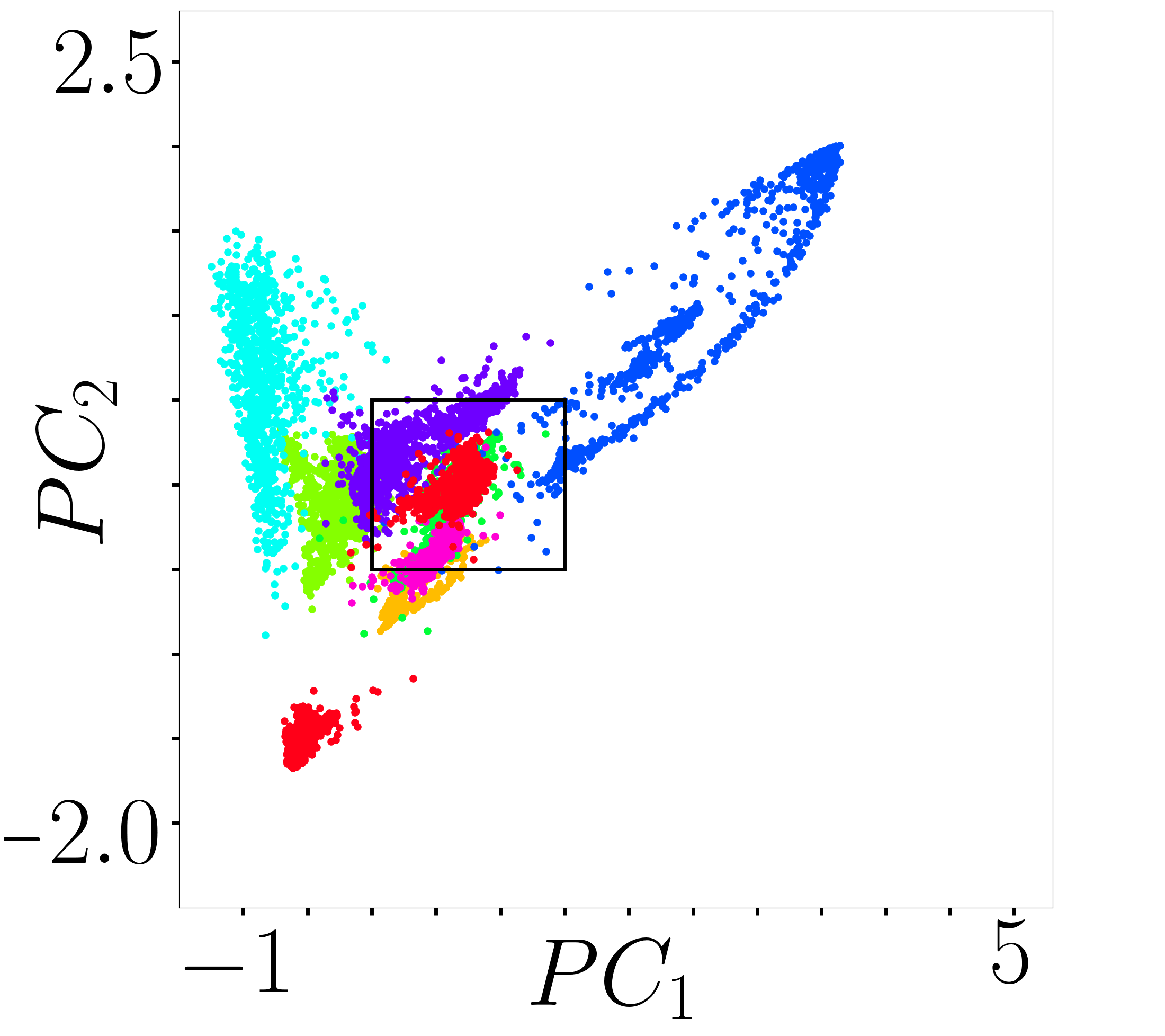}&
  \includegraphics[width=\mywidth\columnwidth]{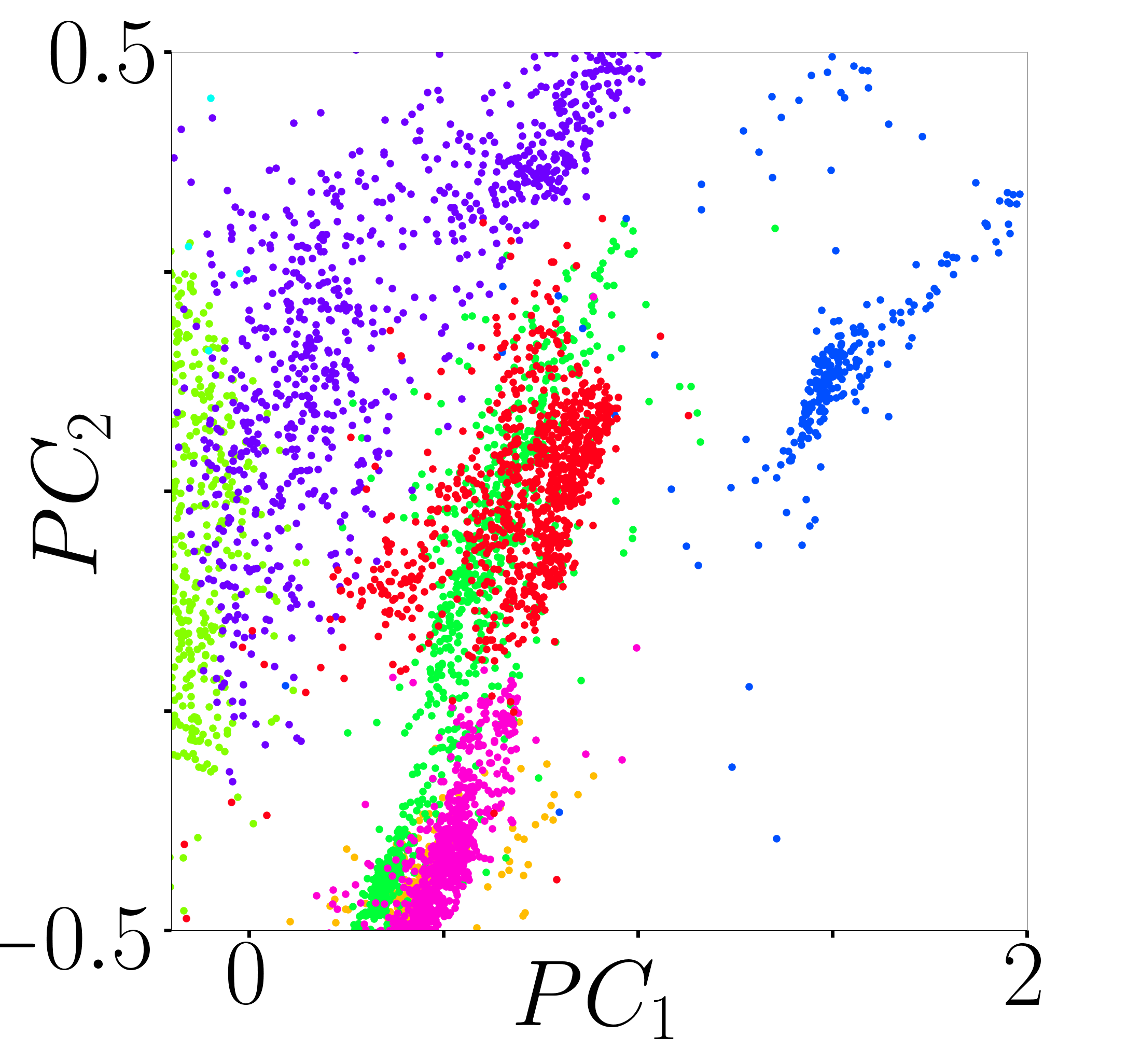}&
  \includegraphics[width=\mywidth\columnwidth]{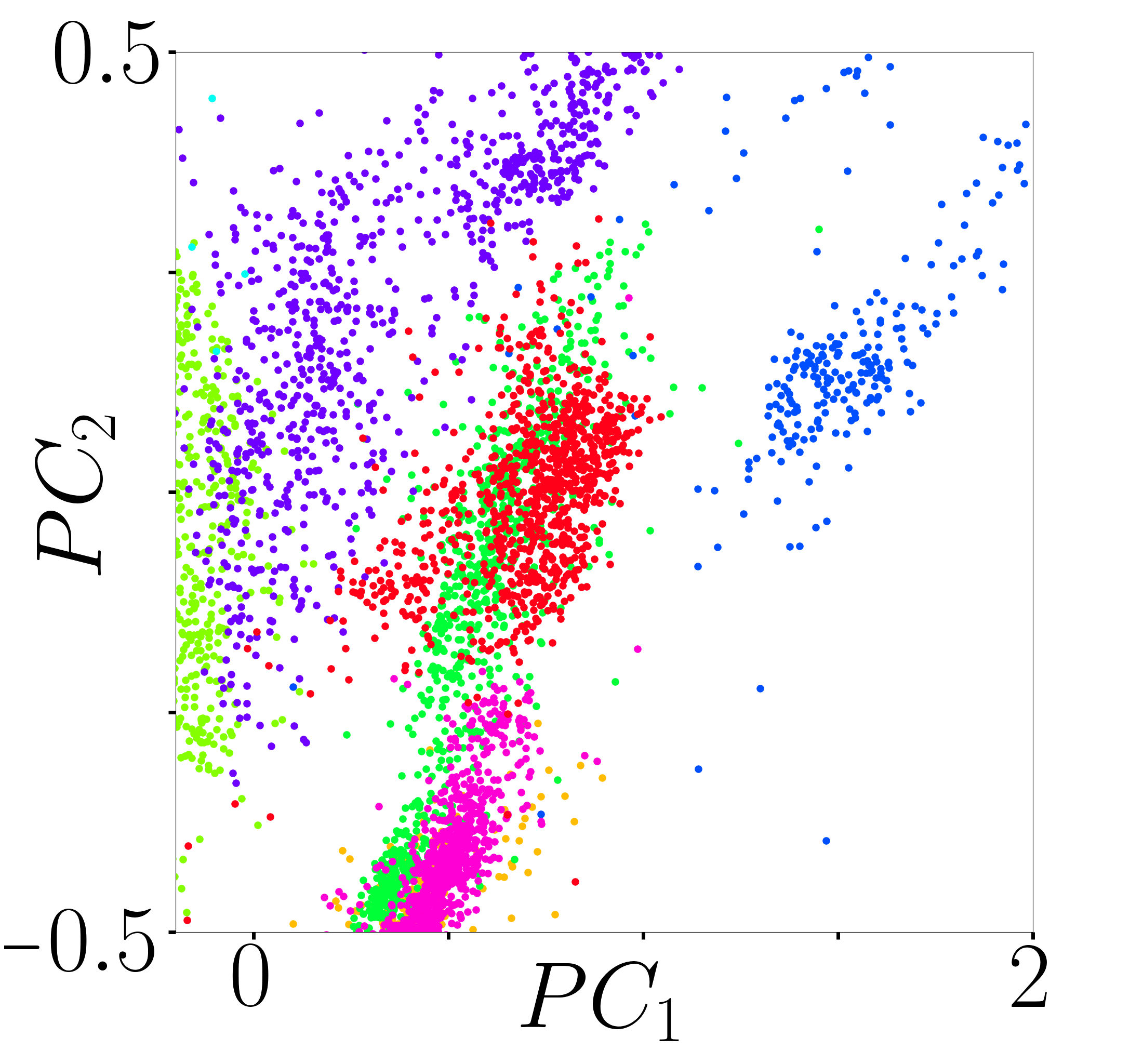}\\
\end{tabular}
	\caption{In our PCA-based augmentation, we randomly shift the values of the original samples along the first principal component (PC) to synthesize new data. A part (black rectangle) of (a)~Pavia University data distribution (after PCA) is zoomed (b), and the synthesized examples are shown in (c) with random $\alpha$ values (for each sample), where $\alpha_{\rm min}=0.9$ and $\alpha_{\rm max}=1.1$ (note that too large $\alpha$'s could adversely impact separability of the classes).}
	\label{fig:pca}
\end{figure}

The first step of our PCA-based data augmentation involves transforming all training samples using PCA (trained over $\TrainingSet$). Afterwards, the first principal component\footnote{However, more principal components could be exploited here.} $PC_1$ (of each sample) is multiplied by a random value $\alpha$ drawn from a uniform distribution $\mathcal{U}(\alpha_{\rm min},\alpha_{\rm max})$, where $\alpha_{\rm min}$ and $\alpha_{\rm max}$ are the hyper-parameters of our method ($\alpha$ is drawn independently for all original examples). This process is visualized in Fig.~\ref{fig:pca}---we can observe that the synthesized examples (Fig.~\ref{fig:pca}c) preserve the original data distribution (Fig.~\ref{fig:pca}b) projected onto a reduced feature space, and preserve inter-class relationships. Finally, these samples are projected back onto the original space (using all principal components to ensure correct mapping), and they are added to the augmented $\TrainingSet$ (if executed offline). This PCA-based augmentation can be applied in both offline and online settings (in both cases, PCA is trained over the original $\TrainingSet$).

\section{Experiments}\label{sec:experiments}

The experimental objective was to verify the impact of data augmentation on the deep model generalization. For online augmentation, we applied our PCA augmentation (PCA-ON), and noise injection (Noise-ON)~\cite{Slavkovikj:2015:HIC:2733373.2806306}, whereas for the offline setting, we used our PCA-based method (PCA), generative adversarial nets (GAN)~\cite{audebert2018generative}, and noise injection (Noise)~\cite{Slavkovikj:2015:HIC:2733373.2806306}. GAN cannot be used online, since it does not modify an incoming example, but rather synthesizes samples which follow an approximated distribution. Finally, we combined online and offline augmentation in PCA/PCA-ON (PCA augmentation is used to both augment the set beforehand, and to generate new samples at test time), and GAN/PCA-ON. For each offline technique, we at most doubled the number of original samples (unless that number would exceed the most numerous class---in such case, we augment only by the missing difference). For online augmentation, we synthesize $\TestTimeGeneratedSamples=4$ samples, and for PCA and PCA-ON, we had $\alpha_{\rm min}=0.9$ and $\alpha_{\rm max}=1.1$.

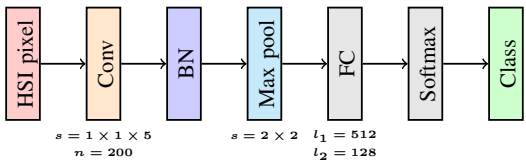
\begin{figure}[ht!]
        \centering
        \resizebox{0.79\columnwidth}{!}{%
            \begin{tikzpicture}[scale=0.8,
                    input/.style={rectangle,draw=black,fill=red!20,inner sep=5pt,minimum height=50pt,minimum width=10pt,text width=5pt,text badly centered,thick},
                    output/.style={rectangle,draw=black,fill=green!20,inner sep=5pt,minimum height=50pt,minimum width=10pt,text width=5pt,text badly centered,thick},
                    convolution/.style={rectangle,draw=black,fill=orange!20,inner sep=5pt,minimum height=50pt,minimum width=10pt,text width=5pt,text badly centered,thick},
                    maxpooling/.style={rectangle,draw=black,fill=cyan!20,inner sep=5pt,minimum height=50pt,minimum width=10pt,text width=5pt,text badly centered,thick},
                    batchNorm/.style={rectangle,draw=black,fill=blue!20,inner sep=5pt,minimum height=50pt,minimum width=10pt,text width=5pt,text badly centered,thick},
                    fullyconnected/.style={rectangle,draw=black,fill=gray!20,inner sep=5pt,minimum height=50pt,minimum width=10pt,text width=5pt,text badly centered,thick},
                    softmax/.style={rectangle,draw=black,fill=gray!20,inner sep=5pt,minimum height=50pt,minimum width=10pt,text width=5pt,text badly centered,thick},
                    myarrow/.style={thick},
                    dottedarrow/.style={dotted, thick}]

                \newcommand{\HSIpixel}{{\rotatebox{90}{{HSI pixel}}}}
                \newcommand{\class}{{\rotatebox{90}{{Class}}}}
                \newcommand{\sep}{20}
                \newcommand{\bigSep}{20}
                \newcommand{\convolution}{{\rotatebox{90}{{Conv}}}}
                \newcommand{\maxpooling}{{\rotatebox{90}{{Max pool}}}}
                \newcommand{\batchnormalization}{{\rotatebox{90}{{BN}}}}
                \newcommand{\deconvolution}{{\rotatebox{90}{{Deconv}}}}
                \newcommand{\fullyconnected}{{\rotatebox{90}{{FC}}}}
                \newcommand{\softmax}{{\rotatebox{90}{{Softmax}}}}
                \newcommand{\convDescrSep}{1}
                \mathchardef\mhyphen="2D 

                \node (in) [input] {$\HSIpixel$};
                \node (conv1) [convolution, right=\sep pt of in]{\convolution};
                \node (convDescription) at ($(conv1.south) - (0 pt, \convDescrSep pt)$)[anchor=north] {{\tiny$\bm{s=1\times 1\times 5}$}};
                \node (convDescription2) at ($(convDescription.south) + (0 pt, 5 pt)$)[anchor=north] {{\tiny$\bm{n=200}$}};
                \node (batchnorm1) [batchNorm, right=\sep pt of conv1]{\batchnormalization};
                \node (maxpooling1) [maxpooling, right=\sep pt of batchnorm1]{\maxpooling};
                \node (convDescription3) at ($(maxpooling1.south) - (0 pt, \convDescrSep pt)$)[anchor=north] {{\tiny$\bm{s=2\times 2}$}};
                \node (fc1) [fullyconnected, right=\sep pt of maxpooling1]{\fullyconnected};
                \node (convDescription3) at ($(fc1.south) - (0 pt, \convDescrSep pt)$)[anchor=north] {{\tiny$\bm{l_1=512}$}};
                \node (convDescription4) at ($(convDescription3.south) + (0 pt, 5 pt)$)[anchor=north] {{\tiny$\bm{l_2=128}$}};
                \node (softmax1) [softmax, right=\sep pt of fc1]{\softmax};
                \node (out) [output, right=\sep pt of softmax1]{\class};

                \draw[->,myarrow] (in) -- (conv1);
                \draw[->,myarrow] (conv1) -- (batchnorm1);
                \draw[->,myarrow] (batchnorm1) -- (maxpooling1);
                \draw[->,myarrow] (maxpooling1) -- (fc1);
                \draw[->,myarrow] (fc1) -- (softmax1);
                \draw[->,myarrow] (softmax1) -- (out);
            \end{tikzpicture}
        } \vspace*{-0.2cm}\caption{Our CNN with $n$ kernels in the convolutional layer ($s$ is a stride) and $l_1$ and $l_2$ neurons in the fully-connected (FC) layers. BN is batch normalization.} \label{fig:1d_network}
    \end{figure}

We exploit our shallow (thus resource-frugal) 1D spectral CNN (Fig.~\ref{fig:1d_network}) coded in \texttt{Python 3.6}~\cite{Nalepa2019GRSL}. Larger-capacity CNNs require longer training and infer slower, hence are less likely to be deployed for Earth observation, especially on board of a satellite. The training (ADAM, learn. rate of $10^{-4}$, $\beta_1 = 0.9$, and $\beta_2 = 0.999$) stops, if after 15 epochs the validation set $\ValidationSet$ (random subset of $\TrainingSet$) accuracy plateaus.

We train and validate the deep models using: (1)~balanced $\TrainingSet$ sets with random pixels (B), (2)~imbalanced $\TrainingSet$ sets with random pixels (IB), and (3)~our patched sets (P)~\cite{Nalepa2019GRSL} (for fair comparison, the numbers of pixels in $\TrainingSet$ and $\TestSet$ for B and IB are close to those reported in~\cite{Gao_2018}). We also report the results obtained using a spectral-spatial CNN (3D-CNN)~\cite{Gao_2018}, trained over the original $\TrainingSet$ (3D-CNN---in contrast to our CNN---does suffer from the training-test information leak problem, and the 3D-CNN results over B and IB are over-optimistic~\cite{Nalepa2019GRSL}). For each fold in (3), we repeat the experiments $5\times$, and for (1)~and (2), we perform Monte-Carlo cross-validation with the same number of runs (e.g., if 5 folds are run $5\times$, we execute $25$ Monte-Carlo runs for B and IB). We report per-class, average (AA), and overall accuracy (OA), averaged across all runs.

We focused on three HSI benchmarks (see their class-distribution characteristics at: \url{http://www.ehu.eus/ccwintco/index.php/Hyperspectral_Remote_Sensing_Scenes}): Salinas Valley, USA ($217\times 512$ pixels, NASA AVIRIS sensor; $\left|\left|\TrainingSet\right|\right|=4320$, $\left|\left|\ValidationSet\right|\right|=480$, $\left|\left|\TestSet\right|\right|=49329$) presents different sorts of vegetation (16 classes, 224 bands, 3.7 m spatial resolution). Indian Pines ($145\times 145$, AVIRIS; $\left|\left|\TrainingSet\right|\right|=2444$, $\left|\left|\ValidationSet\right|\right|=271$, $\left|\left|\TestSet\right|\right|=7534$) covers the North-Western Indiana, USA (agriculture and forest, 16 classes, 200 channels, 20~m). Pavia University ($340\times 610$, ROSIS; $\left|\left|\TrainingSet\right|\right|=2025$, $\left|\left|\ValidationSet\right|\right|=225$, $\left|\left|\TestSet\right|\right|=40526$) was captured over Lombardy, Italy (urban scenery, 9 classes, 103 channels, 1.3 m).

\begin{table*}[ht!]
	\renewcommand{\tabcolsep}{0.13cm}
	\centering
	\scriptsize
		\caption{Per-class, overall (OA), and average (AA) accuracy (in \%) for all scenarios (averaged across all run) for Salinas Valley (16 classes), Indian Pines (16 classes), and Pavia University (9 classes). The darker the cell is, the higher accuracy was obtained).}
		\label{tab:salinas_results}
\begin{adjustbox}{width=\textwidth, center=\textwidth}
	\vrule\pgfplotstabletypeset[%
		 color cells={min=65,max=95,textcolor=black},
		/pgfplots/colormap={blackwhite}{ rgb255=(255,255,255) rgb255=(0, 191, 255)},
		/pgf/number format/fixed,
		/pgf/number format/precision=3,
		col sep=comma,
        every nth row={9}{
            before row=\Xhline{2\arrayrulewidth}
        },
        columns/Set/.style={reset styles,string type, assign cell content/.code={
            \ifnum\pgfplotstablerow=0
            \pgfkeyssetvalue{/pgfplots/table/@cell content}
            {\multirow{9}{*}{##1}}%
            \else
                \ifnum\pgfplotstablerow=9
                \pgfkeyssetvalue{/pgfplots/table/@cell content}
                {\multirow{9}{*}{##1}}%
                \else
                    \ifnum\pgfplotstablerow=18
                    \pgfkeyssetvalue{/pgfplots/table/@cell content}
                    {\multirow{9}{*}{##1}}%
                    \else
                        \ifnum\pgfplotstablerow=27
                        \pgfkeyssetvalue{/pgfplots/table/@cell content}
                        {\multirow{9}{*}{##1}}%
                        \else
                            \ifnum\pgfplotstablerow=36
                        \pgfkeyssetvalue{/pgfplots/table/@cell content}
                            {\multirow{9}{*}{##1}}%
                            \else
                                \ifnum\pgfplotstablerow=45
                                \pgfkeyssetvalue{/pgfplots/table/@cell content}
                                {\multirow{9}{*}{##1}}%
                                \else
                                    \ifnum\pgfplotstablerow=54
                                    \pgfkeyssetvalue{/pgfplots/table/@cell content}
                                    {\multirow{9}{*}{##1}}%
                                    \else
                                        \ifnum\pgfplotstablerow=63
                                        \pgfkeyssetvalue{/pgfplots/table/@cell content}
                                        {\multirow{9}{*}{##1}}%
                                        \else
                                            \ifnum\pgfplotstablerow=72
                                            \pgfkeyssetvalue{/pgfplots/table/@cell content}
                                            {\multirow{9}{*}{##1}}%
                                            \else
                                            \pgfkeyssetvalue{/pgfplots/table/@cell content}{}%
                                            \fi
                                        \fi
                                    \fi
                                \fi
                            \fi
                        \fi
                    \fi
                \fi
            \fi
            }},
		columns/Augmentation/.style={reset styles,string type},
		columns/Fold/.style={reset styles,string type},
		columns/Bands/.style={reset styles,string type},
        every head row/.style={before row=\Xhline{2\arrayrulewidth}, after row=\hline},
        every last row/.style={after row=\Xhline{2\arrayrulewidth}},
        every column/.code={
            \ifnum\pgfplotstablecol=0
                \pgfkeysalso{column type/.add={}{}}%
            \fi
            \ifnum\pgfplotstablecol=1
                \pgfkeysalso{column type/.add={|}{}}%
            \fi
            \ifnum\pgfplotstablecol=2
                \pgfkeysalso{column type/.add={|}{}}%
            \fi
            \ifnum\pgfplotstablecol=18
                \pgfkeysalso{column type/.add={|}{}}%
            \fi
        },
	]{
		Set,Augmentation,C1,C2,C3,C4,C5,C6,C7,C8,C9,C10,C11,C12,C13,C14,C15,C16,OA,AA
		\rotatebox{90}{\textbf{Salinas Valley} (B)},Without,93.74,94.42,85.00,98.47,81.19,99.53,98.98,59.05,92.49,85.37,90.87,88.94,91.90,92.71,63.33,93.93,80.87,88.12
		,Noise,98.60,96.75,95.67,99.25,90.24,99.69,98.96,46.74,97.68,91.18,87.58,98.92,98.38,94.57,79.33,98.77,83.05,92.02
		,GAN,99.36,99.80,99.28,99.09,97.10,99.70,99.45,77.71,99.00,89.29,99.48,99.02,99.84,95.32,72.53,99.14,89.98,95.32
        ,PCA,99.34,99.56,98.38,99.11,96.34,99.79,99.65,76.98,99.36,92.36,96.82,99.59,98.72,98.11,68.68,99.11,89.44,95.12
		,Noise-ON,99.41,98.35,97.22,98.93,90.90,99.81,99.44,75.55,98.95,90.91,91.22,98.54,98.72,95.21,58.67,98.48,86.98,93.14
		,PCA-ON,98.89,99.09,94.99,99.22,95.50,99.76,99.61,68.28,98.46,90.03,93.85,98.58,98.59,96.42,72.74,98.81,87.51,93.93
		,PCA/PCA-ON,99.68,99.15,96.60,99.44,96.41,99.77,99.70,74.73,99.30,92.84,95.76,99.59,98.89,97.70,70.85,99.03,89.18,94.96
		,GAN/PCA-ON,99.12,97.19,95.34,99.09,96.62,99.84,99.57,72.22,98.56,91.88,93.72,98.05,98.75,97.46,68.89,98.57,87.9,94.05
        ,3D-CNN~\cite{Gao_2018}*,99.84,99.30,94.68,99.91,97.51,99.98,99.71,82.82,99.17,96.71,99.42,99.79,99.75,99.73,82.83,99.46,93.04,96.91
		\rotatebox{90}{\textbf{Salinas Valley} (IB)},Without,93.78,96.13,76.46,97.68,81.95,99.34,99.21,83.13,97.55,81,68.01,93.89,97.72,88.28,43.14,89.93,83.57,86.7
		,Noise,98.62,99.32,94.67,95.35,87.46,99.77,99.43,68.61,97.62,83.09,88.26,98.89,98.62,90.47,66.49,97.58,86.02,91.52
		,GAN,97.36,99.75,98.18,96.2,93.88,99.81,99.75,86.26,99.41,93.66,94.5,99.87,98.41,94.84,65.24,98.36,91.11,94.72
        ,PCA,94.02,99.79,93.96,99.21,95.43,99.79,99.63,83.87,99.54,91.87,92.06,99.67,97.46,96.2,64.21,97.96,90.19,94.04
		,Noise-ON,98.5,99.02,94.63,98.66,91.14,99.75,99.4,89.75,99.67,89.36,82.67,97.71,98.11,93.36,45.91,97.51,88.4,92.2
		,PCA-ON,97.83,98.95,91.68,99.16,95.15,99.8,99.57,86.36,99.23,90.82,88.27,99,98.09,95.18,59.25,98.34,89.83,93.54
		,PCA/PCA-ON,99.17,99.33,95.73,99.02,95.09,99.8,99.63,81.81,99.6,92.93,92.11,99.62,97.51,95.68,66.32,98.17,90.32,94.47
		,GAN/PCA-ON,98.01,99.45,94.18,98.13,95.11,99.8,99.71,86.08,99.54,92.24,90.36,98.36,98.06,95.81,65.2,98.5,90.84,94.28
        ,3D-CNN~\cite{Gao_2018}*,99.17,99.21,91.52,99.58,97.37,99.97,99.73,90.03,99.66,96.28,96.46,99.52,99.55,99.00,80.63,96.52,94.27,96.51
        \rotatebox{90}{\textbf{Salinas Valley} (P)},Without,94.06,88.63,49.49,79.23,62.05,79.79,79.52,71.93,89.43,86.07,72.31,85.62,95.71,86.96,53.67,38.18,73.28,75.79
        ,Noise,90.78,92.24,43.94,78.28,66.08,79.77,79.40,66.10,84.98,86.38,72.56,82.31,93.10,86.71,60.09,38.59,72.51,75.08
        ,GAN,96.32,86.44,50.64,79.79,69.36,79.73,79.67,78.05,94.42,85.95,72.83,90.13,96.89,85.60,54.71,43.84,76.09,77.77
        ,PCA,93.76,87.95,48.09,79.56,62.34,79.78,79.63,76.48,91.99,87.33,71.84,96.55,97.90,88.05,58.69,45.23,75.97,77.82
        ,Noise-ON,94.69,92.76,47.91,79.44,61.32,79.83,79.58,84.05,91.16,85.39,69.82,87.48,95.68,87.68,42.29,46.27,75.17,76.58
        ,PCA-ON,90.13,89.33,48.49,79.62,65.73,79.71,79.62,76.60,91.69,87.11,68.04,92.08,98.18,87.69,56.79,45.00,75.59,77.24
        ,PCA/PCA-ON,95.88,90.06,47.10,79.71,66.76,79.84,79.62,78.14,94.24,87.65,73.19,90.75,98.35,87.57,57.01,46.11,76.67,78.25
        ,GAN/PCA-ON,96.02,86.10,50.76,79.73,68.45,79.73,79.67,78.02,94.03,85.72,70.95,89.08,96.93,85.54,54.66,43.87,75.87,77.45
        ,3D-CNN~\cite{Gao_2018},96.49,75.15,39.89,61.61,51.78,79.21,76.81,74.84,78.14,85.69,71.56,76.49,80.86,62.15,61.80,33.00,69.72,69.09
        \rotatebox{90}{\textbf{Indian Pines} (B)},Without,57.89,67.21,70.01,79.55,85.77,95.35,82.89,97.99,70.53,81.13,60.92,79.78,98.28,95.61,34.57,94.61,73.7,78.26
		,Noise,79.38,66.3,73.55,77.59,90.09,97.01,84.38,95.99,74,86.4,62.08,77.76,98.09,91.4,34.82,94.77,74.22,80.22
		,GAN,61.88,65.66,73.83,81.44,91.52,96.07,82.5,97.24,81,76.64,63.69,80.66,97.73,93.19,40.98,93.26,74.34,79.83
        ,PCA,68.75,72.98,76.66,82.36,92.1,95.35,82.5,97.98,83,77.54,70.03,87.39,98.82,93.94,45.27,93.6,78.28,82.39
        ,Noise-ON,66.88,62.44,75.06,76.15,92.04,94.41,79.38,96.34,70,71.08,54.1,75.6,97.45,85.49,30.21,93.84,68.67,76.28
		,PCA-ON,68.44,71.53,74.06,82.24,90.99,95.68,78.13,97.5,72,78.37,59.88,86.34,98.36,95.06,39.61,93.72,74.77,80.12
		,PCA/PCA-ON,79.69,67.84,77.42,83.16,89.29,96.01,89.38,96.45,81,79.8,62.2,84.58,98.18,93.13,48.38,94.53,75.3,82.56
		,GAN/PCA-ON,73.13,66.43,74.2,86.72,93.97,96.5,91.88,97.89,88,75.44,60.93,83.51,97.55,91.33,42.23,95.35,73.74,82.19
        ,3D-CNN~\cite{Gao_2018}*,84.06,67.34,84.62,95.98,93.07,98.66,98.75,98.90,100.00,88.91,70.59,77.13,99.55,91.36,61.12,99.77,79.68,88.11
		\rotatebox{90}{\textbf{Indian Pines} (IB)},Without,38.35,69.36,63.48,56.19,81.64,92.04,66.17,97.68,43,68.29,82.29,62.92,94.95,97.11,52.93,91.32,77.98,72.36
        ,Noise,51.57,62.48,62.64,44.98,73.57,92.35,62.89,92.83,35.69,62.92,68.63,58.56,93.94,94.41,47.3,91.26,69.89,68.5
        ,GAN,37.76,64.85,62.59,55.47,81.81,93.45,63.4,95.82,40.53,64.23,82.16,64.12,91.8,94.61,50.34,88.99,76.47,70.75
        ,PCA,45.55,77.32,69.75,63.88,86.59,94.83,67.44,96.36,48.83,73.83,77.82,76.44,95.36,94.65,56.72,92.06,80.33,76.09
        ,Noise-ON,45.74,67.78,60.05,53.48,84.79,93.35,64.51,97.16,45.14,67.42,79.08,55.77,91.61,92.32,51.75,92.68,75.73,71.41
        ,PCA-ON,43.6,75.34,66.55,54.08,85.13,94.95,72.77,97.23,47.62,71.63,81.01,70.22,92.63,93.88,59.3,92.15,79.67,74.88
        ,PCA/PCA-ON,46.14,75.07,75.27,60.5,85.94,95.21,73.82,97.63,45.91,78.95,78.78,73.32,96.41,94.73,55.97,90.26,80.94,76.49
        ,GAN/PCA-ON,45.79,67.36,63.23,60.47,83.15,94.27,63.25,96.85,48.99,64.34,82.69,65.64,91.28,93.05,53.9,90.61,77.41,72.8
        ,3D-CNN~\cite{Gao_2018}*,62.81,74.81,80.18,83.89,87.67,98.33,76.25,98.53,91.71,84.10,88.85,70.50,99.15,93.92,75.98,99.44,85.90,85.38
        \rotatebox{90}{\textbf{Indian Pines} (P)},Without,17.68,56.89,51.55,36.27,69.02,92.35,0.00,86.95,19.55,60.05,74.05,43.71,94.15,91.18,43.39,45.04,67.11,55.12
        ,Noise,13.39,59.51,52.85,34.82,65.84,91.90,0.00,88.55,17.29,61.27,63.85,40.03,88.14,94.07,40.97,44.78,65.08,53.58
        ,GAN,14.29,62.51,52.24,36.71,69.13,87.83,24.20,68.46,28.32,62.93,67.88,48.32,89.19,76.23,35.18,44.49,66.19,54.24
        ,PCA,15.71,62.13,60.20,42.74,69.51,91.41,0.00,84.29,24.17,64.85,72.00,47.99,93.78,91.08,48.00,44.49,68.84,57.02
        ,Noise-ON,19.11,64.88,56.99,31.81,72.97,88.12,23.20,67.54,25.12,62.13,62.98,49.38,87.52,75.75,35.21,44.35,65.44,54.19
        ,PCA-ON,15.71,59.03,56.99,41.81,66.79,91.94,23.40,70.20,30.93,64.33,70.87,50.79,93.68,75.86,36.09,44.91,67.59,55.83
        ,PCA/PCA-ON,16.07,67.14,55.85,43.00,70.20,93.10,23.94,70.63,33.18,64.66,68.90,52.15,91.96,79.32,34.43,44.45,68.57,56.81
        ,GAN/PCA-ON,14.64,62.21,52.04,36.53,69.02,87.80,23.80,68.12,27.44,62.67,67.44,48.21,89.10,76.19,35.20,44.49,65.97,54.06
        ,3D-CNN~\cite{Gao_2018},5.00,33.70,28.30,17.88,51.32,60.18,0.00,65.99,1.67,53.06,54.27,23.20,65.87,77.01,37.95,37.94,48.89,38.33
        \rotatebox{90}{\textbf{Pavia University} (B)},Without,90.73,86.96,83.09,96.51,99.58,89.59,88.51,80.64,99.86,,,,,,,,88.42,90.61
		,Noise,85.05,84.02,73.69,92.84,99.78,89.95,93.7,88.01,99.91,,,,,,,,86.32,89.66
		,GAN,88.69,94.76,81.11,91.61,99.62,88.32,94.15,85.24,99.94,,,,,,,,91.6,91.49
        ,PCA,89.17,87.52,80.15,96.91,99.67,92.75,91.59,81.47,99.93,,,,,,,,88.85,91.02
        ,Noise-ON,89.26,86.83,68.4,93.23,99.76,65.74,41.1,44.74,99.87,,,,,,,,80.12,76.55
		,PCA-ON,89.07,88.72,78.89,96.36,99.72,88.39,90.53,84.27,99.92,,,,,,,,88.98,90.65
		,PCA/PCA-ON,87.87,91.93,78.97,95.98,99.69,88.84,92.59,82.58,99.87,,,,,,,,90.19,90.92
		,GAN/PCA-ON,88.1,86.86,82.79,95.55,99.72,90.67,94.18,81.5,99.95,,,,,,,,88.24,91.04
        ,3D-CNN~\cite{Gao_2018}*,90.22,89.74,90.72,98.53,99.99,87.04,93.55,88.36,99.91,,,,,,,,90.59,93.12
        \rotatebox{90}{\textbf{Pavia University} (IB)},Without,93.23,97.43,68.83,87.17,99.4,71.84,64.15,81.35,99.65,,,,,,,,89.32,84.78
		,Noise,94.74,93.64,72.67,91.88,99.51,91.9,70.58,83.16,99.84,,,,,,,,71.58,88.66
		,GAN,91.26,97.87,70.31,86.53,99.26,79.89,80.25,87.3,99.44,,,,,,,,91.18,88.01
        ,PCA,94.68,95.86,75.1,92.81,99.49,87.26,73.09,80.42,99.5,,,,,,,,91.59,88.69
        ,Noise-ON,93.92,96.06,62.37,88.26,99.48,60.62,25.43,62.2,99.61,,,,,,,,84.42,76.44
		,PCA-ON,94.02,97.48,74.04,88.47,99.42,80.79,71.17,81.33,99.5,,,,,,,,91.08,87.36
		,PCA/PCA-ON,93.56,95.64,71.17,93.3,99.48,86.43,72.25,84.64,99.7,,,,,,,,91.41,88.46
		,GAN/PCA-ON,92.13,96.35,74.33,89.27,99.48,82.3,73.23,82.51,99.59,,,,,,,,90.72,87.69
        ,3D-CNN~\cite{Gao_2018}*,94.12,98.17,77.00,97.48,99.96,83.69,82.88,87.84,99.58,,,,,,,,93.47,91.19
        \rotatebox{90}{\textbf{Pavia University} (P)},Without,93.40,86.20,47.58,86.89,59.81,27.14,0.00,78.46,79.27,,,,,,,,73.26,62.08
        ,Noise,92.63,84.52,43.76,90.89,59.85,26.31,0.00,77.54,79.72,,,,,,,,72.34,61.69
        ,GAN,92.85,81.91,47.85,91.54,59.83,28.91,0.00,75.23,79.53,,,,,,,,71.55,61.96
        ,PCA,92.89,84.64,47.09,92.86,59.89,29.60,0.00,76.91,79.54,,,,,,,,73.06,62.60
        ,Noise-ON,91.06,85.76,51.40,82.39,59.90,18.77,0.00,60.83,79.75,,,,,,,,70.11,58.87
        ,PCA-ON,93.50,86.16,47.28,84.91,59.88,27.05,0.00,73.38,78.93,,,,,,,,72.63,61.23
        ,PCA/PCA-ON,93.42,86.52,46.88,92.21,59.74,27.68,0.00,78.32,79.60,,,,,,,,73.84,62.71
        ,GAN/PCA-ON,92.87,81.83,47.51,91.53,59.83,28.90,0.00,74.86,79.53,,,,,,,,71.46,62.83
        ,3D-CNN~\cite{Gao_2018},90.66,81.85,41.92,93.02,59.79,25.20,0.00,70.18,79.03,,,,,,,,70.07,60.18
	}\vrule
\end{adjustbox}
{\centering \scriptsize \textbf{\\[0.1cm]}*Note that 3D-CNN gives over-optimistic segmentation results for B and IB due to the training-test information leak problem. For details, see~\cite{Nalepa2019GRSL}.}
\end{table*}

The results (over the test sets $\TestSet$) obtained for the Salinas Valley, Indian Pines, and Pavia University datasets are gathered in Table~\ref{tab:salinas_results}. Introducing augmented samples (in both online and offline settings) helped boost generalization abilities of the deep models in the majority of cases (even up to more than 8\% of OA for GAN, PCA, and PCA/PCA-ON in Salinas, B; only the Noise offline augmentation deteriorated both OA and AA for P). Interestingly, exploring the local neighborhood randomly (Noise and Noise-ON) can notably deteriorate OA and AA. It usually occurs for under-represented classes (e.g.,~C7 in Pavia) since their examples lay close to other-class examples in the discovered feature space (therefore, they can be easily ``confused'' with each other). This problem is addressed by the data-distribution analysis in our PCA-based augmentations. Coupling offline and online augmentation (PCA/PCA-ON and GAN/PCA-ON) gave consistent high-quality results over all sets and all training-test splits, and dealt well with the HSI imbalance (in P, we did not ensure that examples of all classes are included in the original $\TrainingSet$, thus P is very challenging~\cite{Nalepa2019GRSL}).

\begin{table}[ht!]
	\scriptsize
	\centering
	\caption{The results of Wilcoxon's tests (we boldfaced the entries significant at $p<0.05$). Upper triangular of (I) corresponds to B, lower triangular of (I) to IB, and (II) to P.}
	\label{tab:wilcoxon}
	\renewcommand{\tabcolsep}{0.09cm}
	\begin{tabular}{ccllllllll}
\Xhline{2\arrayrulewidth}
&&\multicolumn{1}{c}{(a)}&\multicolumn{1}{c}{(b)}&\multicolumn{1}{c}{(c)}&\multicolumn{1}{c}{(d)}&\multicolumn{1}{c}{(e)}&\multicolumn{1}{c}{(f)}&\multicolumn{1}{c}{(g)}&\multicolumn{1}{c}{(h)}\\
\hline
\multirow{8}{*}{(I)}&(a)&\multicolumn{1}{c}{---}&\textbf{$<$0.05}&\textbf{$<$0.001}&\textbf{$<$0.001}&$>$0.2&\textbf{$<$0.001}&\textbf{$<$0.001}&\textbf{$<$0.001}\\
&(b)&$>$0.2&\multicolumn{1}{c}{---}&$>$0.05&\textbf{$<$0.005}&{$>$0.05}&$>$0.2&\textbf{$<$0.001}&\textbf{$<$0.02}\\
&(c)&$>$0.05&\textbf{$<$0.02}&\multicolumn{1}{c}{---}&$>$0.05&\textbf{$<$0.001}&{$>$0.2}&$>$0.2&$>$0.2\\
&(d)&\textbf{$<$0.001}&\textbf{$<$0.001}&\textbf{$<$0.05}&\multicolumn{1}{c}{---}&\textbf{$<$0.001}&\textbf{$<$0.001}&$>$0.2&$>$0.1\\
&(e)&$>$0.2&$>$0.2&$>$0.2&\textbf{$<$0.001}&\multicolumn{1}{c}{---}&\textbf{$<$0.001}&\textbf{$<$0.001}&\textbf{$<$0.001}\\
&(f)&\textbf{$<$0.001}&\textbf{$<$0.01}&$>$0.1&\textbf{$<$0.05}&\textbf{$<$0.001}&\multicolumn{1}{c}{---}&\textbf{$<$0.02}&{$>$0.2}\\
&(g)&\textbf{$<$0.01}&\textbf{$<$0.001}&\textbf{$<$0.01}&$>$0.2&\textbf{$<$0.001}&\textbf{$<$0.02}&\multicolumn{1}{c}{---}&$>$0.05\\
&(h)&\textbf{$<$0.01}&\textbf{$<$0.02}&$>$0.1&\textbf{$<$0.05}&\textbf{$<$0.01}&{$>$0.2}&\textbf{$<$0.01}&\multicolumn{1}{c}{---}\\
\hline
\multirow{7}{*}{(II)}&(a)&\multicolumn{1}{c}{---}&\textbf{$>$0.05}&$>$0.2&\textbf{$<$0.05}&$>$0.2&$>$0.2&\textbf{$<$0.02}&$>$0.2\\
&(b)&&&\textbf{$>$0.05}&\textbf{$<$0.001}&$>$0.2&\textbf{$<$0.05}&\textbf{$<$0.001}&$>$0.1\\
&(c)&&&&\textbf{$<$0.05}&$>$0.1&$>$0.2&\textbf{$<$0.001}&\textbf{$<$0.001}\\
&(d)&&&&&$>$0.05&$>$0.05&$>$0.2&\textbf{$<$0.02}\\
&(e)&&&&&&\textbf{$<$0.05}&\textbf{$<$0.005}&$>$0.2\\
&(f)&&&&&&&\textbf{$<$0.005}&$>$0.2\\
&(g)&&&&&&&&\textbf{$<$0.001}\\
\Xhline{2\arrayrulewidth}
\multicolumn{10}{c}{{\scriptsize (a)~Without, (b)~Noise, (c)~GAN, (d)~PCA, (e)~Noise-ON, (f)~PCA-ON,}}\\
\multicolumn{10}{c}{{\scriptsize (g)~PCA/PCA-ON, (h)~GAN/PCA-ON.}}
	\end{tabular}
\end{table}

To verify the statistical significance of the results (and see if the differences in the average per-class accuracy are important in the statistical sense), we executed two-tailed Wilcoxon's tests for each dataset split (B, IB, and P) over per-class AA for all HSI. The results reported in Table~\ref{tab:wilcoxon} show that applying HSI data augmentation is beneficial in most cases and delivers significant improvements in accuracy. GAN did equally well as e.g.,~PCA, PCA-ON, and our combined PCA/PCA-ON and GAN/PCA-ON for B, as Noise-ON, PCA-ON, and GAN/PCA-ON for IB, and as Noise-ON and PCA-ON for P. It indicates that employing time-consuming and complex deep-learning engines for data augmentation not necessarily brings larger improvements in the performance of the deep models.

Our combined approaches (PCA/PCA-ON and GAN/PCA-ON) were stable and consistently ensured high-quality generalization (as shown in Table~\ref{tab:salinas_results}) of the deep models over all splits. This stability is also manifested in Table~\ref{tab:summary}, where we summarize the results across all sets (although PCA gave the best accuracy for B, the differences between PCA and PCA/PCA-ON and GAN/PCA-ON are not statistically significant). We can appreciate that our PCA-based augmentation (offline, online, or combined) allowed us to obtain the best generalization---very intuitive PCA-based data-distribution analysis for synthesizing samples outperformed or worked on par with GAN in the case of difficult (small and imbalanced) sets. Finally, our CNN surpassed the accuracy elaborated using a significantly larger 3D-CNN from the literature (with a bigger capacity) for P (note that the results obtained using 3D-CNN for B and IB are over-optimistic due to the intrinsic training-test information leak problem, hence they cannot be considered reliable~\cite{Gao_2018}).

\begin{table}[ht!]
	\scriptsize
	\centering
	\caption{Overal (OA) and average (AA) accuracy averaged across all datasets for all data splits.}
	\label{tab:summary}
	\renewcommand{\tabcolsep}{0.25cm}
	\begin{tabular}{r|rr|rr|rr}
		\Xhline{2\arrayrulewidth}
Set$\rightarrow$	& \multicolumn{2}{c|}{B}  &		\multicolumn{2}{c|}{IB} & 		\multicolumn{2}{c}{P}\\	
\hline
Augmentation$\downarrow$&			OA		&AA		&OA		&AA		&OA		&AA		\\
\hline
Without		&81.00	&85.66	&83.62	&81.28	&71.22	&64.33	\\
Noise		&81.20	&87.30	&75.83	&82.89	&69.98	&63.45	\\
GAN			&85.31	&88.88	&86.25	&84.49	&71.28	&64.66	\\
PCA			&\textbf{85.52}	&\textbf{89.51}	&87.37	&86.27	&72.62	&65.81	\\
Noise-ON	&78.59	&81.99	&82.85	&80.02	& 70.24	& 63.21	\\
PCA-ON		&83.75	&88.23	&86.86	&85.26	&71.94	&64.77	\\
PCA/PCA-ON	&84.89	&89.48	&\textbf{87.56}	&\textbf{86.47}	&\textbf{73.03}	&\textbf{65.92}	\\
GAN/PCA-ON	&83.29	&89.09	&86.32	&84.92	&71.10	&64.78	\\
3D-CNN~\cite{Gao_2018}	&87.77*&	92.71*&	91.21*&	91.03*& 62.89	& 55.87\\
\Xhline{2\arrayrulewidth}
\multicolumn{7}{c}{{\scriptsize *Over-optimistic due to training-test information leak; see~\cite{Nalepa2019GRSL}.}}
	\end{tabular}
\end{table}

To gain better insights into the augmentation performance (and its potential overhead imposed on the deep models in terms of training and/or test times), we collected the average execution times of the most important steps of the investigated methods in Table~\ref{tab:times}. It can be observed that training of GANs is very time-consuming (it was executed using NVIDIA GeForce GTX 1060), and is of orders of magnitude higher than pre-processing in other offline techniques (PCA and Noise). Although all offline augmentations affect the training time of deep networks, these differences are not dramatic. Finally, the online augmentation allowed us to classify test pixels in real-time (note that we report the inference time in ms).

\begin{table}[ht!]
	\scriptsize
	\centering
	\caption{The execution times: (a)~offline augmentation (in s), (b)~CNN training~(s), (c)~classifying one example from $\TestSet$ (ms).}
	\label{tab:times}
	\renewcommand{\tabcolsep}{0.047cm}
	\begin{tabular}{rr|rrr|rrr|rrr}
		\Xhline{2\arrayrulewidth}
 & Set$\rightarrow$ & \multicolumn{3}{c|}{B} &  \multicolumn{3}{c|}{IB} &  \multicolumn{3}{c}{P}\\
 \hline

& Aug.$\downarrow$ & Sa & IP & PU & Sa & IP & PU & Sa & IP & PU\\
\hline
\multirow{3}{*}{(a)}&Noise		&0.14	&0.02		&0.05		&0.10		&0.06		&0.03		&0.08		&0.05		&0.03		\\
&GAN							&529.10	&2048.47	&241.18		&555.77		&1938.28	&617.40		&594.28		&960.14		&1151.71	\\
&PCA							&0.10	&0.05		&0.02		&0.09		&0.06		&0.02		&0.08		&0.05		&0.02		\\
\hline
\multirow{4}{*}{(b)}&Without	&103.15	&64.01		&14.97		&116.40		&63.51		&16.05		&118.98		&64.46		&26.08		\\
&Noise							&176.20	&56.91		&32.60		&127.31		&76.85		&28.38		&104.44		&67.21		&31.47		\\
&GAN							&146.21	&146.22		&48.11		&192.36		&92.87		&29.33		&102.40		&67.97		&25.32		\\
&PCA							&167.83	&91.14		&26.14		&156.30		&84.28		&23.53		&183.89		&81.13		&27.40		\\
\hline
\multirow{6}{*}{(c)}&Without	&0.09	&0.11		&0.07		&0.09		&0.11		&0.07		&0.10		&0.15		&0.09		\\
&Noise							&0.09	&0.11		&0.07		&0.09		&0.11		&0.07		&0.09		&0.15		&0.09		\\
&GAN							&0.09	&0.11		&0.07		&0.09		&0.12		&0.07		&0.10		&0.14		&0.10		\\
&PCA							&0.09	&0.11		&0.07		&0.09		&0.11		&0.07		&0.10		&0.15		&0.09		\\
&Noise-ON						&1.65	&1.90		&1.46		&1.75		&1.84		&1.44		&2.03		&2.23		&1.79		\\
&PCA-ON							&1.97	&1.96		&1.61		&2.00		&1.99		&1.55		&2.18		&2.40		&2.01		\\

\Xhline{2\arrayrulewidth}
 \multicolumn{11}{c}{{\scriptsize Sa---Salinas Valley, IP---Indian Pines, PU---Pavia University.}}
	\end{tabular}
\end{table}

\section{Conclusions}\label{sec:conclusions}

In this letter, we introduced a new online HSI data augmentation approach which synthesizes examples at test time. It is in contrast to other state-of-the-art hyperspectral data augmentation techniques that work offline (i.e.,~before the deep-network training to increase the training set cardinality and representativeness). Our experimental study, performed over three HSI benchmark sets (with different training-test data splits) and coupled with statistical tests revealed that our online augmentation is very flexible (different augmentations can be applied here), improves the generalization abilities of deep neural networks, and works in real-time. Also, we showed that combining online and offline augmentation leads to consistently well-performing models. Finally, we proposed a principal component analysis based augmentation which operates extremely fast, synthesizes high-quality data, outperforms other augmentations for small and imbalanced sets, and is applicable in online and offline settings.

\ifCLASSOPTIONcaptionsoff
  \newpage
\fi

\bibliographystyle{ieeetran}
\bibliography{IEEEabrv,ref_all}

\end{document}